\documentclass[10pt,twocolumn,letterpaper]{article}

\usepackage[pagenumbers]{cvpr} 

\usepackage{graphicx}
\usepackage{amsmath}
\usepackage{amssymb}
\usepackage{booktabs}
\usepackage{color}

\usepackage[pagebackref,breaklinks,colorlinks]{hyperref}

\usepackage[capitalize]{cleveref}
\crefname{section}{Sec.}{Secs.}
\Crefname{section}{Section}{Sections}
\Crefname{table}{Table}{Tables}
\crefname{table}{Tab.}{Tabs.}

\def\cvprPaperID{9} 
\def\confName{CVPR}
\def\confYear{2023}

\begin{document}

\title{CVB: A Video Dataset of Cattle Visual Behaviors}

\author{Ali Zia*$^{1,2}$~
Renuka Sharma*$^{2}$~
Reza Arablouei$^{2}$~
Greg Bishop-Hurley$^{2}$~
Jody McNally$^{2}$\\
Neil Bagnall$^{2}$~
Vivien Rolland$^{2}$~
Brano Kusy$^{2}$~
Lars Petersson$^{2}$~
Aaron Ingham$^{2}$\\
The Australian National University,Australia$^1$\\
CSIRO, Australia$^2$\\
{\tt\small \{Ali.Zia,Renuka.Sharma,Reza.Arablouei,Greg.Bishop-Hurley,Jody.McNally,}\\
{\tt\small	Neil.Bagnall,Vivien.Rolland,Brano.Kusy,Lars.Petersson,Aaron.Ingham\}@csiro.au}\\
}
\maketitle
\newcommand\blfootnote[1]{%
	\begingroup
	\renewcommand\thefootnote{}\footnote{#1}%
	\addtocounter{footnote}{-1}%
	\endgroup
}
\blfootnote{* Equal contribution}


\begin{abstract} 
%


Existing image/video datasets for cattle behavior recognition are mostly small, lack well-defined labels, or are collected in unrealistic controlled environments. This limits the utility of machine learning (ML) models learned from them. Therefore, we introduce a new dataset, called Cattle Visual Behaviors (CVB), that consists of 502 video clips, each fifteen seconds long, captured in natural lighting conditions, and annotated with eleven visually perceptible behaviors of grazing cattle. We use the Computer Vision Annotation Tool (CVAT) to collect our annotations. To make the procedure more efficient, we perform an initial detection and tracking of cattle in the videos using appropriate pre-trained models. The results are corrected by domain experts along with cattle behavior labeling in CVAT. The pre-hoc detection and tracking step significantly reduces the manual annotation time and effort. Moreover, we convert CVB to the atomic visual action (AVA) format and train and evaluate the popular SlowFast action recognition model on it. The associated preliminary results confirm that we can localize the cattle and recognize their frequently occurring behaviors with confidence. By creating and sharing CVB\footnote{\href{https://doi.org/10.25919/3g3t-p068}{The Cattle Visual Behaviors (CVB) dataset.}}, our aim is to develop improved models capable of recognizing all important behaviors accurately and to assist other researchers and practitioners in developing and evaluating new ML models for cattle behavior classification using video data.
\end{abstract}


\section{Introduction}
\label{sec:intro}



In 2020, the global livestock industry was worth about USD 1.5 trillion and the global beef industry alone was valued at approximately USD 300 billion. Accurate knowledge of cattle behavior is essential for monitoring their welfare, health, and productivity, and has a range of applications in animal science and livestock management. For instance, behavioral information can be used to surveil the feeding and grazing habits of cattle and identify any anomalies that may indicate health issues or stress. Moreover, comprehensive cattle behavioral data can help optimize livestock management practices by, for example, identifying the most efficient grazing patterns or stocking rates for different types of pasture or cattle. 
This can also reduce negative environmental effects of cattle farming such as soil erosion or water pollution. In addition, by improving health and productivity of cattle, farmers can increase their profitability and contribute to the sustainable growth of the livestock industry.

\begin{figure*}[ht]
\centerline{\includegraphics[width=.95\textwidth]{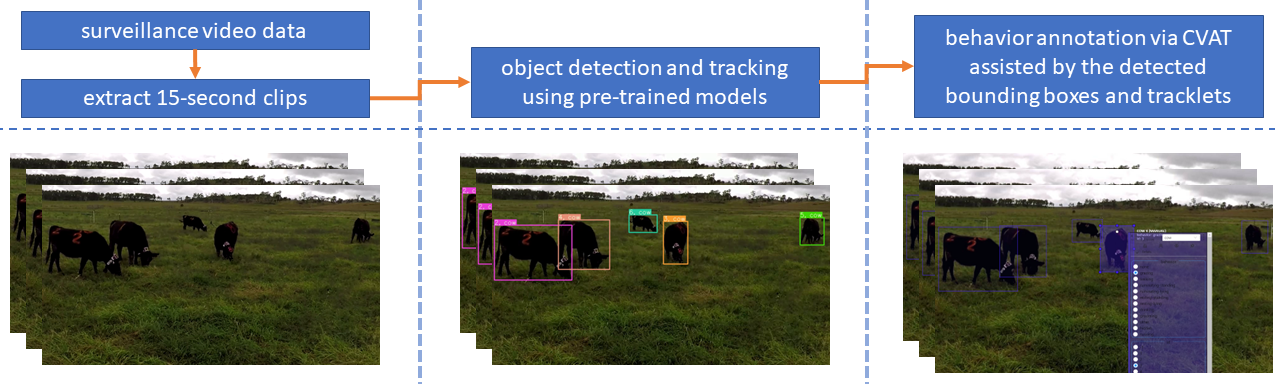}}
\caption{\footnotesize  \textbf{Overall annotation process utilizing CVAT and pre-trained object detection and tracking models.}}
\label{fig:annotation_framework}
\end{figure*}


Behavior (or action) recognition from video data is an emerging field of study in machine learning (ML)~\cite{kleanthous2022survey, ravbar2019automatic, broome2023going, zuerl2022automated, nguyen2021video, wang2022multi}. Previous research~\cite{Jinkun2019} shows that sufficiently large and exhaustive datasets are necessary for developing robust ML models. Most existing video datasets for cattle behavior recognition are limited in multiple ways. 
(i) They are often small in size or lack comprehensive annotations,
which limits the usefulness of the models learned from them~\cite{bybee2001introduction}. 
(ii) They are collected in controlled environments~\cite{van2013automated} (e.g., artificially-lit laboratory or indoor conditions), which makes the learned models hard to generalize or transfer to different contexts (e.g., a different habitat or natural lighting conditions).
(iii) They contain few annotations~\cite{Ng_2022_CVPR} (e.g., few behaviors) making them unsuitable for any in-depth analysis of animal behavior. 



To achieve robust cattle behavior recognition from video data, several steps are taken including the detection of individual animals in the scene, identifying common and unique characteristics of animals, tracking each animal over time, and classifying each animal's behavior over the course of the video. 
Identifying individual cattle is challenging as they may lack distinctive features or be heavily occluded. Changing illumination/lighting, camera angles, or background can pose additional challenges. 


To tackle some of the above-mentioned challenges, we collected a video dataset in natural lighting conditions using high-resolution cameras placed at four corners of an experimental field containing eight Angus beef cattle. To curate our dataset, we have divided the recorded videos into sections of fixed time spans, applied pre-trained object detection and tracking models to localize the cattle in each frame, and used domain experts and a video annotation tool to assign a behavior label to each cattle in each frame.
In this paper, we introduce this dataset, which we call Cattle Visual Behaviors (CVB). It consists of 502 15-second video clips annotated with eleven visually perceptible cattle behaviors on a small pasture. We evaluate the performance of a popular action recognition algorithm on CVB. 
The results show that we can localize the cattle and recognize the most frequently occurring behaviors with good accuracy. Our objective is to utilize CVB to develop models that can recognize less common cattle behaviors in a real-world setting.
We hope that CVB can also help other researchers and practitioners to develop and evaluate new ML models for animal behavior classification using video data.




\section{Related work}
Many computer vision (CV) researchers have attempted to create ML models and tools for recognizing and analyzing animal behavior~\cite{Ng_2022_CVPR,liang2018benchmark}. However, most existing relevant animal behavior datasets are small, fragmented, or environment-specific. 
In addition, these datasets typically contain still images rather than videos, are specific to certain environments, or only pertain to the classification of animal species. A few examples are iNaturalist~\cite{van2018inaturalist}, Animals with Attributes~\cite{xian2018zero}, Caltech-UCSD Birds~\cite{wah2011caltech}, and Florida Wildlife Animal Trap~\cite{gagne2021florida}).

Datasets like AwA-Pose~\cite{banik2021novel}, AcinoSet~\cite{joska2021acinoset}, Animals-10~\cite{narayan2022adjusting}, Animal Kingdom~\cite{Ng_2022_CVPR} consider the behaviors of animals in their associated environmental settings. Most of them focus on a small number of behavior classes for a particular animal species/genus while they may contain multiple types of animals. In CVB, we concentrate on only one animal species, i.e., cattle (\emph{Bos taurus}), and attempt to annotate all related visually perceptible behaviors.

The two most relevant datasets to ours are Animal Kingdom~\cite{Ng_2022_CVPR} and Dataset on Large Animals~\cite{liang2018benchmark}. The former covers a diverse set of animals consisting of mammals, reptiles, amphibians, birds, fishes, and insects while the latter consists of videos of 60 large animals including cattle. Our CVB dataset contains 502 labeled videos, each with 450 frames annotated by domain experts. All cattle in every frame of every video is labeled with its associated behavior and assigned a unique identification number, which remains consistent across the video. CVB encompasses a thorough observation of cattle behavior as well as their visual localization and tracking. We believe this is a significant contribution towards cattle behavior recognition from video data as datasets like CVB can ultimately lead to ML algorithms that automatically recognize and track cattle behavior and hence substantially reduce the effort required by domain experts to annotate video data and subsequently other sensor data.

There are various CV-based approaches related to cattle identification and behavior recognition~\cite{oliveira2021review}, including visual localization and individual identification~\cite{andrew2017visual}, detecting cow structure~\cite{liu2020video}, keypoint detection and MAP estimation~\cite{t2020long}, livestock monitoring with transformers~\cite{TangiralaBLGTA21}, instance segmentation using Mask R-CNN~\cite{salau2020instance}, cattle counting using Mask R-CNN~\cite{xu2020automated}, cattle gait tracking~\cite{gardenier2018object}, detection of dairy cows~\cite{tassinari2021computer}, RGB-D video-based individual identification~\cite{okura2019rgb}, part segmentation using keypoint guidance~\cite{naha2021part}, video-based cattle identification~\cite{nguyen2021video}, and 3D feature aggregation~\cite{ma2021voxelized}. Some approaches that incorporate interactions between the cattle and their rearing environment include CV-based indexes for analyzing the response to rearing environment~\cite{massari2022computer}, tracking and analyzing social interactions between cattle~\cite{ren2021tracking}, and CV and deep learning based approaches for behavior recognition in pigs and cattle~\cite{chen2021behaviour}. A survey on cattle re-identification and tracking~\cite{ravoor2020deep} covers various approaches to re-identification and tracking of cattle that can uniquely identify cattle across all the video frames and track their behaviors.

Some approaches originally developed for human action recognition~\cite{gu2018ava} such as SlowFast~\cite{feichtenhofer2019slowfast} can also be extended to the application of animal behavior recognition~\cite{wishart2018hmdb,soomro2012ucf101,carreira2017quo,gu2018ava}. SlowFast is an efficient yet effective method that processes semantic and motion information simultaneously using two parallel pathways operating at different (slow and fast) frame rates.

\section{Data Collection}

Annotating animal behavior is not straightforward, especially for non-experts, because many behaviors appear similar. In addition, the transition between behaviors is not always distinct, and 
when animals are in close proximity, there may be significant occlusions. 
Long videos require substantial resources to generate a comprehensive set of annotations. 
The raw videos of our CVB dataset were captured using GoPro5 Black cameras placed at the four corners of a 25mx25m area in natural lighting conditions at a resolution of 1920x1080 pixels and 30 frames per second (FPS). We create the dataset by annotating 502 sections of the raw data with the location and behavior of each cattle in each frame. Each section is 15 seconds long. We obtain initial annotations for cattle localization by applying a pre-trained YOLOv7 object detection model~\cite{wang2022yolov7} to all video frames. Domain experts adjust the detected bounding boxes and assign cattle behavior labels to them using the Computer Vision Annotation Tool (CVAT)~\cite{boris_sekachev_2020_4009388}. 

To cope with occlusions, we track the movements of individual animals. We track the uniquely identified cattle over 450 frames of each 15-second-long video and obtain tracklets corresponding to each unique animal and behavior instance. The tracklets also facilitate the annotation by capturing the motion of each cattle across the frames and obviating the need to manually assign a behavior to each cattle in every frame.

\subsection{Cattle and their Behaviors}
 
We consider eleven visually perceptible and mutually exclusive cattle behaviors in CVB. They are
grazing, walking, running, ruminating-standing, ruminating-lying, resting-standing, resting-lying, drinking, grooming, other (when the behavior is not any of the previous ones), and hidden.
The considered behaviors exhibited by cattle are fundamental to grazing cattle, specifically beef cattle, as they spend a significant part of their lives performing them. Moreover, these behaviors are crucial in evaluating and monitoring their productivity, feed efficiency, energetic dynamics, health, welfare, and regulatory compliance. For instance, the knowledge of the duration and timing of a cattle's grazing is critical in determining its herbage dry matter intake from the pasture~\cite{smith21}. Understanding when and for how long a cattle ruminates or rests can also provide insight into the animal's health and well-being status~\cite{schirmann12}. Analyzing the duration of rumination versus grazing can offer valuable insights into the quality of the pasture or feed. Monitoring walking behavior can help measure the animal's energy expenditure and discover its movement patterns, while identifying drinking behavior is vital in ensuring the cattle's access to water and compliance with relevant regulations. In addition, as grooming is important for the health and well-being of cattle (e.g., by helping maintain hygiene), understanding the related behavioral patterns and habits can be informative. 

\subsection{Object Detection}
 
Initial object detection via a pre-trained YOLOv7 model allows the domain experts to annotate the videos significantly faster.
YOLOv7 is an advanced version of the popular YOLO object detection algorithm that uses a single convolutional neural network (CNN) to identify objects in real-time video streams. YOLOv7 achieves high accuracy in object detection, with mean average precision (mAP) scores that are comparable to or better than other state-of-the-art object detection algorithms. This is due to a number of innovations, such as feature fusion to combine information from different layers of the CNN and spatial attention mechanisms to emphasize on relevant regions of the image. In addition, YOLOv7 can generally run in real time, with a processing speed of up to 90 (FPS) on a single GPU. This makes it well-suited to our application, which requires fast and accurate object detection.


\subsection{Tracklets}
A tracklet captures the movement of a cattle exhibiting a behavior that may commence at any frame within a 450-frame video and conclude either before or at the last frame. We use the Botsort algorithm~\cite{aharon2022bot}, which is a state-of-the-art tracking approach, to track the movement of animals in each video stream by creating a trajectory for each animal based on its position in successive video frames. Botsort employs a set of rules to predict any object's movement, including its velocity, acceleration, and movement direction, and subsequently updates its position in the following frame. In order to predict the movement of an object, Botsort takes into account its previous position and the positions of the nearby objects. It also employs some heuristics to adjust the predicted trajectory in response to abrupt changes in the object's movement, such as a change in speed or direction.
Furthermore, Botsort incorporates camera motion compensation to predict the correct location of the object which may fail due to camera motion in complex scenarios.
Botsort can track multiple objects simultaneously, even if they move in close proximity. This makes it particularly useful for tracking social interactions between animals. It also handles occlusions to some extent by predicting the location of the occluded object based on its trajectory.

 \section{Main Characteristics of CVB}
 
After cattle detection and tracking by applying relevant pre-trained models, domain experts did review, verify, and adjust the detected bounding boxes and tracklets as required. They also added the behavior labels.

The CVB dataset contains eleven cattle behaviors, ranging from frequent behaviors such as grazing and resting to infrequent behaviors such as drinking and grooming. Cattle may exhibit these behaviors with different frequencies and durations at different conditions or times. 
Some behaviors occur much less frequently than others, which makes them difficult to observe. In CVB, grazing is the most frequent behavior, occurring $39\%$ of the time, followed by resting-lying ($15\%$), resting-standing ($12\%$), hidden ($10\%$), and ruminating-lying ($6\%$). Each of the less frequent behaviors, namely, walking, drinking, ruminating-standing, grooming, running, and other, occurred less than $5\%$ of the time.
%
There is a natural imbalance in the prevalence of the considered behaviors in CVB that pertains to a Zipfian distribution~\cite{bybee2001introduction}. Recognition models need to be designed to perform well on long-tailed behavior distributions~\cite{van2017devil} that reflect real-world scenarios, rather than on artificially balanced datasets.
In addition, CVB includes transitions between different behaviors, such as a cattle transitioning from grazing to drinking or resting to walking. It also contains light or heavy occlusions, variations in lighting conditions, and different camera angles, which further complicate the annotation process.

 \section {Experiments and Analysis}
 \begin{figure}[t]
\centering
\begin{subfigure}[t]{0.49\textwidth}
   \includegraphics[width=1\linewidth]{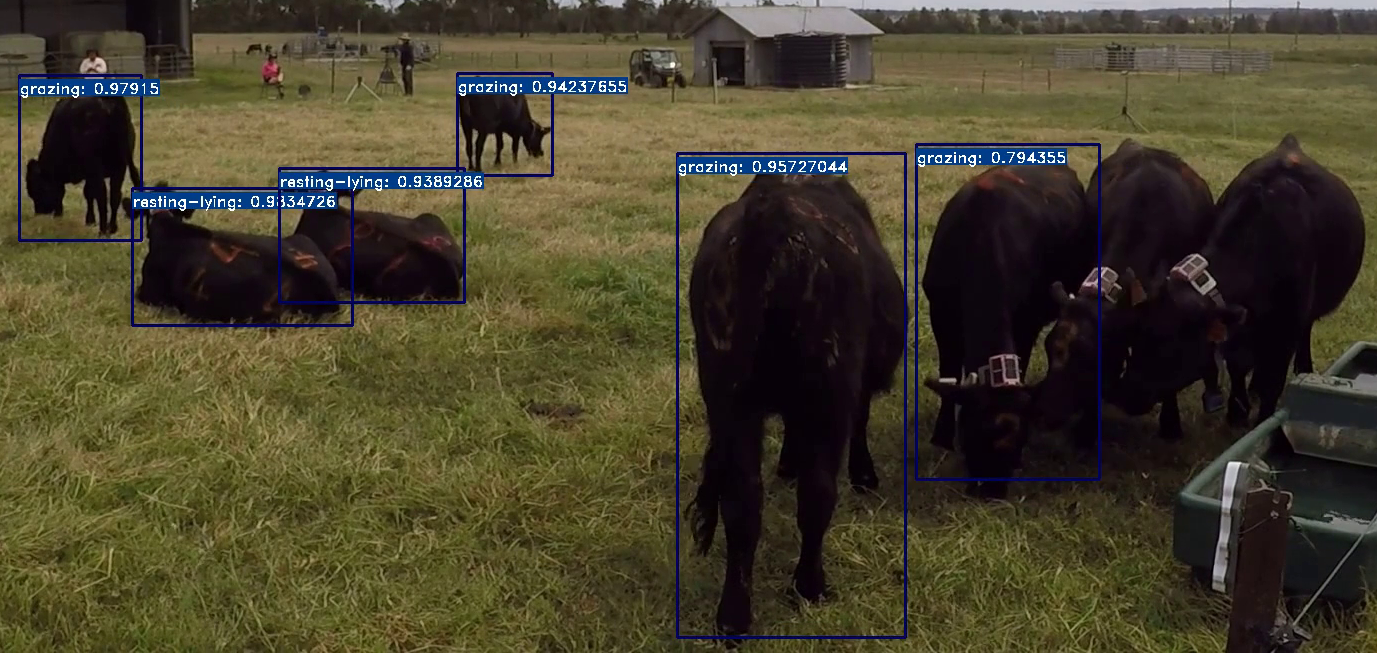}
   \caption{}
   \label{fig:Ng1} 
\end{subfigure}
\begin{subfigure}[t]{0.49\textwidth}
   \includegraphics[width=1\linewidth, height=0.35\linewidth]{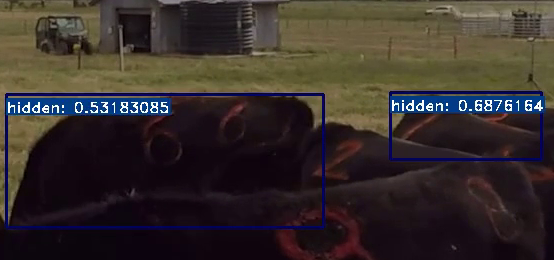}
   \caption{}
   \label{fig:Ng2}
\end{subfigure}

\caption[Inference]{Example inference results using SlowFast~\cite{feichtenhofer2019slowfast}, showing (a) various behaviors in a scene and (b) multiple hidden instances. 
}
\vspace{-15pt}
\label{fig:inference}
\end{figure}
We generate initial cattle localization and tracking annotations using YOLOv7 and Botsort algorithms, respectively. We further refine them manually while annotating the behavior of each cattle using CVAT. The object detector may occasionally mistake a shed or tree in the background for cattle. We remove such erroneous detections during annotation. 
We corrected $136,598$ bounding boxes in $65,300$ frames out of $225,900$ annotated frames. This means only $29\%$ of the detections made by YOLOv7 and Botsort required adjustment, which indicates that the use of these pre-trained models greatly reduced the annotation burden. 


Recognizing the behavior of multiple cattle in natural settings is a complex problem that may involve interactions among cattle or with certain objects present in their surroundings, such as water trough when drinking. AVA~\cite{gu2018ava} dataset format is suitable for studying such scenarios. Therefore, we also provide CVB with annotations that are converted to the AVA format.

We train the SlowFast model on CVB (converted to the AVA format) and evaluate its performance using an 80:20 train/test data split. 
The average test accuracy for grazing is $73.96\%$ when the cattle localization IOU is greater than $50\%$ with an average confidence of $89\%$. Similarly, for resting-lying, the test accuracy is $58\%$.
We present the inference results for a few frames in Fig.~\ref{fig:inference}. Fig.~\ref{fig:inference}(a) shows a scenario where cattle exhibit a variety of behaviors ranging from more frequent behaviors like grazing and resting-lying to less frequent behaviors like mild aggression (performed by two animals on the right). SlowFast can detect the frequent behaviors with good accuracy but has difficulty in recognizing the less frequent behaviors. Fig.~\ref{fig:inference} (b) shows an example when multiple cattle are partially occluded and labeled as hidden in the dataset. As the figure shows, SlowFast has some success in identifying this peripheral class despite the paucity and high ambiguity of the relevant training data.


We continue to extend the CVB dataset and make it publicly available to encourage more researchers to consider tackling the important problem of recognizing animal behavior from video data. We also 
conduct ongoing research on devising new deep feature representations that can lead to efficient methods capable of correctly classifying a wider range of behaviors, including those that are less common.





\section{Conclusion}

We introduced the CVB dataset that consists of 502 videos of eight grazing cattle annotated with eleven visually perceptible cattle behaviors. The dataset contains the location and appropriate behavior label for each cattle in each frame of the videos. We created CVB utilizing pre-trained object detection and tracking models, CVAT, and manual expert labor. 
The evaluation of the performance of a widely-used action recognition algorithm (i.e., SlowFast) trained over CVB indicated that accurate identification of the location of the cattle and recognition of their most commonly observed behaviors is achievable.
Our aspiration is for CVB to assist fellow researchers and practitioners in the creation and assessment of novel ML models that can accurately recognize animal behavior through video data.
{\small
\bibliographystyle{ieee_fullname}
\bibliography{egbib}
}

\end{document}